# Segmentation of 3D pore space from CT images using curvilinear skeleton: application to numerical simulation of microbial decomposition

Olivier MONGA[a,b], Zakaria BELGHALI[b], Mouad KLAI[b], Lucie DRUOTON[c], Dominique MICHELUCCI[c], Valérie POT[d]

[a] IRD, UMMISCO, Unité de Modélisation Mathématique et Informatique des Systèmes Complexes, F-93143, Bondy, France

[b] Cadi Ayyad University, Faculty of Sciences Semlalia, Department of Mathematics, BP 2390, Marrakech

[c] LIB, UFR Sciences et Techniques, University of Burgundy, F-21000 Dijon, France

[d] UMR 1402, ECOSYS, INRA, 78850, France

**Corresponding author:** Olivier Monga, Olivier.monga@ird.fr

**Authors contributions**

**Olivier MONGA**: He is the PhD supervisor of Zakaria BELGHALI and Mouad KLAI, and was the co-supervisor of Lucie DRUOTON ; writing of the paper ; methodology ; implementation of the numerical simulation of microbial decomposition and testing on the datasets.

**Zakaria BELGHALI**: PhD student ; methodology ; implementation of the geometrical modelling method ; test on the datasets ; figures of the paper.

**Mouad KLAI**: PhD student, Matrix explicit numerical scheme, methodology overall diffusive conductance.

**Lucie DRUOTON**: Seed work on the use of curvilinear skeleton for segmenting pore space ; writing of the paper.

**Dominique MICHELUCCI**: He was the co-supervisor of Lucie DRUOTON ; writing of the paper ; methodology.

**Valérie POT:** LBM implementation and test ; writing of the paper.

**Abstract:**

Recent advances in 3D X-ray Computed Tomographic (CT) sensors have stimulated research efforts to unveil the extremely complex micro-scale processes that control the activity of soil microorganisms. Voxel-based description (up to hundreds millions voxels) of the pore space can be extracted, from grey level 3D CT scanner images, by means of simple image processing tools. Classical methods for numerical simulation of biological dynamics using mesh of voxels, such as Lattice Boltzmann Model (LBM), are too much time consuming. Thus, the use of more compact and reliable geometrical representations of pore space can drastically decrease the computational cost of the simulations. Several recent works propose basic analytic volume primitives (e.g. spheres, generalized cylinders, ellipsoids) to define a piece-wise approximation of pore space for numerical simulation of draining, diffusion and microbial decomposition. Such approaches work well but the drawback is that it generates approximation errors.

In the present work, we study another alternative where pore space is described by means of geometrically relevant connected subsets of voxels (regions) computed from the curvilinear skeleton. Indeed, many works use the curvilinear skeleton (3D medial axis) for analyzing and partitioning 3D shapes within various domains (medicine, material sciences, petroleum engineering, etc.) but only a few ones in soil sciences. Within the context of soil sciences, most studies dealing with 3D medial axis focus on the determination of pore throats. Here, we segment pore space using curvilinear skeleton in order to achieve numerical simulation of microbial decomposition (including diffusion processes). We validate simulation outputs by comparison with other methods using different pore space geometrical representations (balls, voxels).

# 1 Introduction

Supported by the recent advances in X-scanner Computed Tomography (CT) sensors, the understanding and modelling of biological and physical dynamics in porous media (soil, rocks, etc.) at microscale is becoming an important emerging research problem. Indeed, CT X-scanners provide nowadays high-resolution (a few micron meters and down to 0.1 micron meter) 3D images of porous media samples. Thus, since a decade, this technological advance allows to tackle computational modelling of porous media dynamics at very fine scales. A key challenge, within this research focus, consists in using efficiently the spatialization information contained in CT images, in order to achieve numerical simulation. Practically, from the rough grey level image data, we extract voxel-based representation of relevant structures such as pore space. Commonly, basic image processing tools (histogram analysis, thresholding…) (Horaud and Monga 1995) can perform this task, ending up with a set of voxels (up to hundreds millions) representing pore space.

Within the context of biological dynamics simulation, the direct use of the primary voxel-based representation leads to algorithms underlying huge computing costs. For instance, classical methods, such as Lattice Boltzmann Model (LBM) and Partial Differential Equations systems solving, are accurate but highly space and time consuming. In order to decrease the computational cost, the use of more compact geometrical representations of pore space appear as valuable alternative. That is one reason why, mostly in the last decade, some efforts have been devoted to the development of advanced computer vision and computational geometry tools in order to represent pore space in a compact way. These tools are aiming to provide compact, intrinsic and relevant geometrical representations from the raw set of voxels (Monga et al. 2007, Ngom et al. 2012, Youssef et al. 2007, Monga 2007, Kemgue et al. 2019, Gong et al. 2020). These advanced data representations can lead to considerable speed up of the numerical simulation algorithms. We point out that such methods take into account the exact real pore space that is different from classical pore network models which define an idealized pore space representation.

This paper deals with the segmentation (partitioning) of pore space into relevant regions using the properties of the curvilinear skeleton. We validate our pore space representation within the context of microbial decomposition including diffusion processes. We represent the result of pore space segmentation with an attributed valuated adjacency graph. Each region (considered as a pore) is attached to a node of the graph. The information attached to each region (volume, center of gravity, etc.) is stored in the corresponding node. Two contiguous regions are linked with an edge in the graph. A contact area, representing the exchange capacity between two regions, is calculated for each pair of neighboring regions. This information and the distance between the gravity centers of the two regions are attached to the corresponding edge. Then, numerical simulations of biological processes are performed from the graph same as in (Monga et al. 2022, Monga et al. 2014, Mbe et al. 2022, Leye et al. 2015). We validate our approach on real data by comparing the results with the one provided by Lattice Boltzmann Method (LBM) and by MOSAIC method (ball based representation of pore space). We use the data sets described in Monga et al. (2014) and Mbe et al. 2022.

The contributions of this paper can be summarized as follows:

- We propose to segment the pore space by means of the simple maximal branches of the curvilinear skeleton. The graph (network) which takes into account all adjacency relationships between regions (or pores) is built.
- We show that our pore space representation (segmentation) can be used for graph-based simulation of microbial decomposition, including diffusion processes. We use an implicit numerical scheme same as in Monga et al. 2022.
- We test our approach on three different sets of data described in Monga et al. (2022) and Monga et al. (2014). The results are compared with the ones provided by LBM (Pot et al. 2014) and the ones provided by MOSAIC (Monga et al. 2022 and Mbe et al. 2022).
- We emphasize that the use of pore space representation by curvilinear skeleton yields significant computing time gain (1/6 to 1/10) without loss of generality.

This paper is split into 6 sections. Section 2 deals with background, definitions, state of the art, and big picture of our work. The real image data sets used to validate this work is presented in Section 3. Section 4 copes with the technical details of our method. It presents the computation of the simple branches of the curvilinear skeleton, the segmentation into regions of the set of voxels, the construction of the adjacency graph of the regions, and graph-based numerical simulations of diffusion and microbial decomposition. Section 5 shows results on different real data sets. We give some comparisons with other alternative works referenced in the literature. Section 6 concludes, giving possible future extensions.

## 2  Background and literature review

### 2.1  Introduction

Since last decade, computational geometry tools are increasingly used within the field of computational soil sciences. In this context, the surface skeleton, and the curvilinear skeleton, appear as key tools to address pore space geometrical modelling. Figure 1 illustrates the notion of surface skeleton and curve skeleton. Some recent works are also aiming at defining analytic representation of pore space using sophisticated primitives. This section reminds related definitions, properties and methods.

### 2.2  Surface skeleton

The surface skeleton of a shape, also close to its medial axis, is the locus of the centers of maximal spheres inside the shape (Boissonnat and Yvinec 1998). By definition, a sphere is maximal if not strictly included in another sphere inside the shape. The union of maximal spheres covers the shape. This notion is widely used in computational geometry since Blum (1967)'s paper. The surface skeleton of a 3D shape is an union of points, curves and surfaces.

One can find in the literature various ways to compute the surface skeleton, depending on the initial description of the shape: triangular mesh, subset of voxels in a 3D image, point cloud (seeds) sampling the boundary of the shape. For instance, regarding shapes represented with a set of voxels in a 3D image, Xia and Tucker (2009) compute a distance map (or distance field) to the shape boundary by solving the Eikonal equation (of light ray trajectory) with the Fast-Marching method. This distance map gives, for each voxel in the shape, the geodesic distance (i.e. the length of the shortest path inside the shape) to the boundary of the shape. The voxels of the surface skeleton are then extracted with the Laplacian of this map. Melkemi (1997) and Amenta et al.(2001) describe the Power Crust algorithm which approaches the medial axis of a point cloud (seeds) sampling the shape boundary. The computation of the medial axis is done through the computation of the Voronoi diagram of the point cloud. The Voronoi diagram is the dual of the Delaunay triangulation, and both are computationally equivalent. The boundary surface of the shape can be reconstructed from the surface skeleton of the point cloud. The surface skeleton can also be approximated using the 3D Delaunay Triangulation of a finite set of points forming a discretization (sampling) of the shape boundary. The set of the centers of the spheres circumscribed to the Delaunay tetrahedra (Delaunay spheres), and included within the shape, gives an approximation of the skeleton (Monga 2007, Ngom et al. 2012, Monga et al. 2007). Theoretically, it can be shown that when the sampling converges uniformly towards the boundary shape, then the set of centers converge toward the surface skeleton (Boissonnat and Yvinec 1998).

Actually, the sensitivity of the surface skeleton to little changes, in the shape boundary, is a drawback for several applications. Therefore, many works cope with more robust definitions of the surface skeleton for addressing various tasks such as animation, motion tracking, shape recognition and analysis... For instance when shape details are irrelevant, the surface skeleton is simplified by means of the hierarchical removal of small maximal spheres. Such strategies are related to the notion of λ-skeleton (Chazal and Lieutier 2005). Anyway, within our specific context, small maximal spheres can be filled with water, and thus diffusion processes can unfold in such pores. Therefore, the surface skeleton should be carefully simplified.

Another limitation of the surface skeleton is that it can be hardly used directly to segment the initial shape. Indeed, connected (2D manifold) subsets lying on a surface can hardly be represented in a compact way, on the contrary (1D manifold) curves in a 3D space can be manipulated in much easier manners. For instance, Reniers and Telea (2008) use the (1D manifold) boundaries of the surface skeleton to segment the shape. Thus, they resort to a 1D (manifold) representation. Indeed, the segmentation of a complex shape is much easier to address by means of the more recent notion of curve skeleton Silver et al. 2007 and Dey et al. 2006.

## 2.3   Curvilinear skeleton

### 2.3.1   Introduction

The curvilinear skeleton is intended to be a 1D (manifold) representation of a 3D shape centered on it, which captures its topology and geometry. With this aim, Silver et al. (2007) and Dey et al. (2006) discussed desired properties of curvilinear skeletons. Several definitions of the curvilinear skeleton have been proposed. For instance the curvilinear skeleton can be a subset of the surface skeleton, fulfilling a supplementary condition, like the singularity of the Hessian of the distance map. Mathematical morphology proposed other definitions, based on morphological operators: erosion (or thinning), homotopic thinning… Several discrete or continuous curve skeleton extraction methods (Bertrand and Couprie 2007, Zwettler et al.

2008 and Wang et al. 2012) have been proposed in the field of Discrete Geometry, Shape Analysis, etc. Bertrand and Couprie 2007 describe the homotopic thinning algorithm. Zwettler et al. 2008 adapt this algorithm in the specific case of tubular surfaces representing blood vessels for medical diagnostics. Dealing with another medical application, Wang et al. 2012 propose a method which iteratively contracts an initial mesh, to get the curve skeleton. Sobiecki et al. 2013, 2014 compare different methods for computing curve skeletons. They give a comparison of voxel-based methods (Sobiecki et al. 2014) with mesh contraction-based methods. Both are controlled with geometric criteria such as homotopy, thickness, preservation of details (Sobiecki et al. 2013). Figure 2 shows the curve skeletons (homotopic thining) of three familiar shapes. Indeed, the curvilinear skeleton is much simpler and more computationally convenient than the surface skeleton: the former is 1D (manifold), the latter is 2D (manifold). Such a 1D (manifold) shape representation is especially convenient for various application. Similarly to the surface skeleton, the curvilinear skeleton faces robustness problems. That is the reason why some works are studying more robust definitions of the curvilinear skeleton (Silver et al. 2007). Anyway, within our particular context, we should keep the whole curvilinear skeleton in order to avoid losing relevant information. The use of curve skeleton for segmenting 3D shapes was explored only recently. Reniers and Telea 2007 segmented the shape according to the critical points of the curve skeleton. The critical points were defined as triple cross road points i.e. belonging to at least three branches of the curve skeleton. The curve skeleton is then segmented using the geodesic distances (length of the shortest paths inside the volume) of the critical points to the boundaries. On the same principle, Brunner and Brunnett 2004 presented a mesh segmentation algorithm combining voxelization and homotopic thinning. In a first step, the mesh, attached to the shape boundaries, is transformed into a volume voxel-based representation. Then the computation of the curvilinear skeleton is achieved by homotopic thinning. The curvilinear skeleton is considered as a graph. Each branch of the graph is associated with a 3D volume region in the Voronoi sense. The final result is deduced by binding to each region the corresponding part of the boundary mesh. These two works Reniers and Telea 2007 and Brunner and Brunnett 2004 considered shapes for which the curve skeleton is relatively simple, that is generally not the case in soil sciences.

### 2.3.2 Pore space geometrical modeling using curvilinear skeleton

In soil sciences, Lindquist and Venkatarangan 1999 and Lindquist et al. 1996 used medial axis (notion close to surface skeleton) in order to analyze spatial distributions of pores in soil. Youssef et al. 2007 address quantitative 3D characterization of the pore space of real rocks within petroleum context. They cope with the segmentation of the pore space from the curve skeleton. The curve skeleton is calculated by homotopic thinning. A minimal radius is attached to each voxel of the curve skeleton. This minimal radius is defined as the geodesic distance to the border of the pore space. Voxels where this radius is a local minimum are removed: each connected component of the remaining skeleton is considered as a pore. Two neighboring pores communicate by a channel. Pores and channels are respectively vertices and edges of a graph, used for scattering simulation. The authors compare the results of the measurements with those of simulations, for various porous media like sands and carbonates. The provided network is directly derived from the segmentation of the curvilinear skeleton. Therefore, some adjacency relationships between neighboring pores are lost that is not the case in the present work.

### 2.3.3 Analytic representation

Another alternative consists in defining intrinsic analytic piece wise approximation of pore space using volume basic primitives. The drawback of most analytic approaches lays on the a priori choice of given geometric primitives. Analytic scheme could be also implemented after a first segmentation of the pore space, for instance using the methods described above. Within pore network modeling context, a well-known algorithm is the Maximal Ball algorithm (Al-Kharusi and Blunt 2007). It consists in computing the maximal spheres contained in the pore space and then approximating the volume by a subset of the spheres. This algorithm is modified by Dong and Blunt 2009, which propose a faster way to determine the set of maximal balls. In their approach, the network representing the pore space is composed of balls and cylinders attached to pores and throats. The biological simulation is then performed using this network. Monga et al. 2007 propose to approximate the pore space with an optimal subset of maximal spheres and cylinders. This optimal set is derived from the minimum set of balls (in cardinal sense) recovering the $\lambda$-skeleton. Afterward, numerical simulation of microbial decomposition is performed using balls network (Monga et al. 2008). Recently, several works proposed using more sophisticated geometrical primitives such as ellipsoids and generalized cylinders (Kemgue et al. 2019, Kemgue and Monga 2018 and Ngom et al. 2012). For instance, the algorithm described in Kemgue et al. 2019 includes two steps. First, maximal spheres are clustered using k-means segmentation method. Afterward, each cluster is approximated with a primitive. Finally, an optimal region growing stage allows to reduce the number of primitives. Regarding air-water interfaces extraction, pore space representation with spheres or ellipsoids gives good results (Kemgue et al. 2019). Within other application contexts, several works are also dealing with complex volume shape modeling using sophisticated primitives like for instance ellipsoids (Mokhtari et al. 2014, Banégas et al. 1999a, 1999b, 2001 and Ngom et al. 2012). For example, in medical context, Banégas et al. 1999a, 1999b, 2001 investigated complex volume shape modeling using spheres and ellipsoids. These approaches can be hardly directly extended to pore space geometrical modelling due to the higher shape complexity. Indeed, in the specific case of pore space modelling, the joint use of computational geometry tools like 3D Delaunay triangulation and skeleton extraction is a key element for the algorithms robustness.

## 3 Description of our method

### 3.1 Principle

From the 3D CT images we extract a voxel based representation of the pore space (see figures 5,6,7). Then, the pore space is represented by a set of voxels typically hundreds millions of voxels. In a first step, we compute the curvilinear skeleton of this set of voxels which is a 3D curve (1D manifold) (Lee et al. 1994). In a second step, we segment the curvilinear skeleton into maximal simple branches. In a third step, we compute for each voxel the simple branch which is the closest in the Euclidean distance sense. Afterward, we attach to each simple branch the corresponding set of voxels. In a fourth step, we label (i.e. segment) the pore space into connected sets of voxels (regions). In a fifth step, we build an attributed adjacency graph of regions from the segmentation where each node is attached to a set of connected voxels corresponding to a pore. Finally, we use this compact pore space representation to achieve numerical simulation of microbial decomposition.

### 3.2 Computation of the curvilinear skeleton

Let S be the set of voxels forming the pore space. Let B be the surface border of S. We assume w.l.o.g. (Without Loss of Generality) that S is connected. If it is not the case, we work separately on each connected component of S. We compute the curvilinear skeleton of S by means of successive erosions using the homotopic thinning algorithm proposed by Kollmannsberger et al. 2017.

As mentionned in Kollmannsberger et al. 2017 and Lee et al. (1994), at each stage of the erosion process, voxel (i,j,k) is deleted if the following conditions hold :

- (i,j,k) lies on B (S boundary), i.e. at least one of its 26 neighbors lies outside S.
- (i,j,k) is not an ending point (an ending point has only one neighbor), regarding the 26-neighborhood.
- (i,j,k) removal does not modify Euler characteristics (topological condition).
- (i,j,k) removal does not modify the number of connected components (topological condition).

Erosion process (thinning) is iterated until the fix point is reached (idempotence). Homotopic thinning algorithm is robust and works for any shape represented by voxels, even for disconnected B with connected S (volume shape with holes). By construction, the curve skeleton thickness is 1 voxel. In general, the curve skeleton of a given shape is unique, but some very symmetric shapes can have several curve skeletons, for instance the cube $[0,9]^3$ minus the concentric cube $[3,6]^3$. For these cases, homotopic thinning algorithm comes up with a curvilinear skeleton depending on implementation details (data reading sense), like the order in which boundary voxels of S (B) are considered. Such cases are unlikely when modelling natural shapes such as in soil sciences.

### 3.3 Segmentation of the curvilinear skeleton into simple branches

We propose to segment the curvilinear skeleton in order to extract the pore network.

The curvilinear skeleton is represented by a graph in which each node is attached to a voxel. Neighboring voxels are linked by an edge, we use the 26-connexity.

In the curvilinear skeleton graph, there are different kinds of nodes (cf. figure 3), depending on their degree in graph theory sense (number of neighbors):

- Ending nodes are nodes with only one neighbor.
- Simple nodes are nodes with exactly two neighbors.
- Interior nodes are nodes with more than two neighbors.

A branch is a maximal set of connected simple nodes. Ending nodes and interior nodes are located at the extremity of branches. Interior nodes lie at the intersection of branches. figure 3 shows an example of skeleton graph. We segment the graph into a set of branches using a straightforward algorithm. Thus, we got a set of branches forming a partition of the curvilinear skeleton.

### 3.4 From curve skeleton branches to pore network graph

Afterward, each voxel of S is associated with the nearest branch of the curve skeleton in the sense of Euclidean distance. We got a partition of S composed of regions forming practically connected set of voxels. In the very seldom cases, where the set of voxels attached to a branch would not form a connected component, we split into connected sets. Another way of tackling this problem would be to use the geodesic distance instead of the Euclidean one as proposed in Druoton et al. 2018. By this way, we partition the pore space into connected set of voxels corresponding each to a single branch of the curvilinear skeleton. The segmentation (partition) of S is directly defined by the nearest sets attached to the branches. Three examples of volume shape segmentation for familiar shapes are given in figure 2. At the end, we get an attributed adjacency graph representing the pore network denoted as G. The graph of regions defining the pore network is built as follows. Each node of the graph is attached to a region (pore) corresponding to a branch of the curve skeleton. Each arc of G is attached to a pair of neighboring regions. The adjacency relationships are computed by building a 3D image L(i,j,k) where each voxel is attached to the label of the region where it is included. In case of (i,j,k) would not belong to S, we set L(I,j,k) to 0. The arcs of G and the area of the contact surfaces between adjacent regions (pores) are determined by means of a one pass scanning. For each voxel (i,j,k) of S we look for the three neighbors: (i+1,j,k), (i,j+1,k), (i,j,k+1). For each neighbor belonging to S, we updated the graph adjacencies (arcs) and the areas of the contact surfaces (arc features) by looking to the values of the label image (L). If the two labels are the same, we do nothing. If the two labels are different we create eventually a new arc in the graph and increment the area of the contact surfaces between the two regions.

The previous algorithm finds all neighboring relationships between regions and all corresponding areas of contact surfaces. It is used because not all edges of G can be created considering only the branches of the curve skeleton. Indeed, some regions may share a common boundary surfel although their branches do not have a common interior node. Neglecting regions adjacencies such as in in Brunner and Brunnett 2004 and Youssef et al. 2007 could be a sensible drawback for many applications. For instance, in the case of diffusion simulation, it would imply that flows would transit only along branches of the curvilinear skeleton. Brunner and Brunnett (2004) did not use the curve skeleton for diffusion simulation and therefore neglected these neighboring relationships. In the same way, Youssef et al. (2007) also ignored some neighborhood relationships. The information given by the graph of the regions is summarized in figure 4.

## 3.5 From pore network graph to microbial decomposition simulation

### 3.5.1 Graph updating to simulate biological dynamics

From the previous stages, we get an attributed relational graph G such that:

- Each node is attached to a connected set of voxels (region, pore) corresponding to a branch of the curvilinear skeleton ; we associate to each node the coordinates of the inertia center of the set of voxels and its volume (number of voxels).

- Each arc is attached to an adjacency relationship between two regions (pores) ; we associate to each arc the area of the contact surface between the two corresponding regions (pores).

Afterward, we use the graph G, representing the pore space, to simulate microbial decomposition. The global scheme of the numerical simulation is the same as the one described in Monga et al. 2022. The difference relies on the geometrical representation of the pore space. In a previous work (Monga et al. 2022), the pore space is described by the minimal set of balls recovering the surface skeleton (Monga et al. 2007). The drawback is that a part of the pore space (typically 15%) is lost due to the piece wise approximation errors. Here our geometrical representation covers completely the set of voxels corresponding to pore space. Thus, for instance, we got exact values for the contact surfaces areas between two pores.
As described through details in (Monga et al. 2022) , we get the following implicit numerical scheme:

We note: $\Theta_{ij} = D_c \frac{s_{ij}}{d_{ij}} \delta t$ where $D_c, s_{ij}, d_{ij}, \delta t$ are respectively the diffusion coefficient, the area of the contact surface between node (region, pore) i and j, the distance between the two inertia center of regions i and j, the discretization step time.

We note $c_i, v_i, N_i, \vartheta(N_i)$, respectively the concentration at node $i$, the volume of node $i$, node $i$, the neighbors of node $N_i$.

The relationship between concentrations at successive iterations can be expressed as follow:

$$\begin{bmatrix} c_1 \\ c_2 \\ \vdots \\ c_p \end{bmatrix}^k = \begin{pmatrix} \frac{1}{v_1} & \cdots & 0 \\ \vdots & \ddots & \vdots \\ 0 & \cdots & \frac{1}{v_p} \end{pmatrix} \begin{pmatrix} v_1 + \sum_{N_j \in \vartheta(N_1)} \Theta_{1,j} & \cdots & -\Theta_{1,p} \\ \vdots & \ddots & \vdots \\ -\Theta_{p,1} & \cdots & v_p + \sum_{N_j \in \vartheta(N_p)} \Theta_{p,j} \end{pmatrix} \begin{bmatrix} c_1 \\ c_2 \\ \vdots \\ c_p \end{bmatrix}^{k+1} \quad (1)$$

We solve the above system thanks to conjugate gradient method (Monga et al. 2022

### 3.5.2 Diffusive overall conductance computation

In the case where the Fick flows are defined between two regions, we must multiply the flows with a coefficient $\propto_{i,j}$ called diffusive overall conductance such as in Niloo et al. (2022). In order to define $\propto_{i,j}$, we calibrate by comparison with the diffusion simulation using balls network and LBM. Practically, we find out that we can set $\propto_{i,j}$ to a constant value α. Therefore the Fick flow becomes:

$$F_{i,j} = \propto_{i,j} \frac{-D_c S_{i,j} \Delta c_{ij} \delta t}{d_{i,j}} \quad (2)$$

The calibration principle is to adjust such that the diffusion using the curvilinear skeleton based pore network description fits with the ones provided by LBM and balls network methods. Practically, we define diffusion benchmarks as described in Monga et al. 2022, and optimize the correlation between the outputs. By this way we find out that setting $\propto_{i,j}$ to the constant value 0.35 allows good fitting results. We show that the above statement applies for different sets of data.

### 3.5.3  Step time validation using explicit numerical scheme

We can check the pertinence of the step time value for the implicit numerical scheme using the explicit numerical scheme. In a previous paper we proposed to implement implicit numerical scheme using directly the graph Monga et al. 2022. In the present work we use a matrix representation yielding substantial computational gain. Thanks to the same calculation scheme than in section 3.5.2 we get:

$$\begin{bmatrix} c_1 \\ c_2 \\ \vdots \\ c_p \end{bmatrix}^{k+1} = \begin{pmatrix} \frac{1}{v_1} & \cdots & 0 \\ \vdots & \ddots & \vdots \\ 0 & \cdots & \frac{1}{v_p} \end{pmatrix} \begin{pmatrix} v_1 - \sum_{N_j \in \vartheta(N_1)} \Theta_{1,j} & \cdots & \Theta_{1,p} \\ \vdots & \ddots & \vdots \\ \Theta_{p,1} & \cdots & v_p - \sum_{N_j \in \vartheta(N_p)} \Theta_{p,j} \end{pmatrix} \begin{bmatrix} c_1 \\ c_2 \\ \vdots \\ c_p \end{bmatrix}^k \quad \textbf{(4)}$$

As stated in Monga et al. 2022, the drawback of the explicit numerical scheme is that it needs very small step times in order to avoid negative values. Thus, we use it only to check the validity of the step time of the implicit numerical scheme.

### 3.5.4  Microbial decomposition simulation

In order to simulate microbial decomposition, we use the scheme described in Monga et al. 2022. The principle is to discretize in time the process and to implement successively transformation and diffusion processes. The diffusion processes are implemented by means of the implicit numerical scheme (see 3.5.2 and 3.5.3). The computational cost is roughly proportional to the number of graph nodes, which is the number of balls in Monga et al. 2022. In most cases, the number of regions provided by the curvilinear skeleton based method is much lower (ratio of 1/10 to 1/20).

## 4   Data sets

This section is devoted to the description of the data sets used to validate our method. We use two different real data sets presented respectively in Monga et al. 2014, in Monga et al. 2022 and Mbe et al 2022.

Dealing with the data set of Monga et al. 2022 and Mbe et al. 2022, The soil used in this study is a sandy loam soil (sand, silt, clay: 71, 19 and 10% soil mass, respectively) from the Bullion Field, which is an experimental site situated at the James Hutton Institute in Invergowrie (Scotland). A detailed description of the soil samples and methods used to produce CT images can be found in Juyal et al. 2018. The sizes of the 3D CT image is $512 \times 512 \times 512$ and the resolution $24\mu m \times 24\mu m \times 24\mu m$. After binarization we got 22720090 voxels which forms 17% of the whole volume.

Regarding the data set of Monga et al. 2014, it consists in quartz sand of Fontainebleau forest. We use two CT images corresponding respectively to high and low porosity. The size of the 3D CT images is $512 \times 512 \times 512$ and the resolution $4.5\mu m \times 4.5\mu m \times 4.5\mu m$.

We validate the whole scheme: geometrical pore space modelling using curvilinear skeleton (see section 2.3.2), diffusion calibration by computing the diffusive overall conductance (see section 3.5.2), microbial decomposition simulation including diffusion process (dissolved organic matter).

# 5 Results

## 5.1 Geometrical modelling

We implement the pore space geometrical modeling scheme described in section 3.4 for the three data sets: sandy loam soil (*dataset 1*), Fontainebleau quartz high porosity *(dataset 2)*, Fontainebleau quartz low porosity (*dataset 3).* Figures 5 to 19 illustrate the different stages of the geometrical modelling process for datasets 1-2-3.

The compacity and the coherence of the regions (pores) shapes show the pertinence of the approach which is confirmed by the next processing stages: diffusion and microbial decomposition numerical simulation. The whole geometrical modelling phase takes roughly 20 minutes CPU time on a regular midrange laptop (AMD Ryzen 5 5500U, 8 Gigabytes RAM). We point out that the geometrical modeling stage has to be ran only once before simulating the various biological dynamics scenarios.

## 5.2 Diffusion simulation

In order to use our geometrical representation for the (numerical) simulation of diffusion processes, we calibrate it by comparison with other approaches. We have implemented the same calibration process than the one described in Monga et al. 2022. Here, the goal of the calibration phase is to define the value of the diffusive overall conductance $\propto_{i,j}$. We notice that for our three datasets we can set $\propto_{i,j}$ to a constant value ($\propto_{i,j}= 0.35$). As in Monga et al. 2022, we put a given mass of organic matter into the first two planes. We compare the simulation results of the diffusion process using the three geometrical representations of the pore space: voxel based (LBM), ball based (MOSAIC), region based (approach of the present paper). For the voxel based approach, Lattice Boltzmann Method was used to simulate diffusion process. For balls and regions based approaches, the graph updating allows to simulate diffusion (see section 3.5). We get good fitting of the region based approach with the other ones by setting $\alpha_{i,j}$ to a constant value ($\alpha_{i,j} = 0.35$). Figure 20 shows the calibration results for dataset 1.

The computational cost of the diffusion phase is roughly proportional to the number of nodes of the geometrical primitives graph. For data set 1, using the geometrical modelling scheme described in Monga et al. (2022), we got 191583 balls describing pore space. Thanks to the curvilinear skeleton based algorithm described here, we got 18508 regions. Thus, the computing time for the diffusion simulation is roughly divided by a ratio of ten. For dataset 3, we got 478191 balls and 72691 regions, yielding a ratio of 7.

## 5.3 Microbial decomposition simulation

This section deals with the validation of our pore space geometrical modeling from curvilinear skeleton for microbial decomposition simulation. The computational scheme used is the same than the one described in Monga et al. 2022. The principle is to process sequentially, with a step time, diffusion process (implicit numerical scheme) and transformation process.

For dataset 1, we compare the results obtained by using the skeleton based pore network graph with the ones provided by LBM and MOSAIC in Monga et al. 2022. We take exactly the same scenario than the one described in Monga et al. 2022 and Mbe et al. 2022. The biological parameters of Monga et al. (2014) from Arthrobacter sp. 9R were taken for DOM degradation with 9.6 $j^{-1}$ for the maximum growth rate, 0.001 gC $g^{-1}$ for the constant of half-saturation, 0.2 $j^{-1}$ for the respiration rate, 0.5 $j^{-1}$ for the mortality rate and 0.55 for the proportion of µO that returns to DOM. The parameters of Iqbal et al (2014) were taken for decomposition with 0.3 $j^{-1}$ for the decomposition rate of POM and 0.001 $j^{-1}$ for the decomposition rate of SOM. We put 5.2 10$^7$ bacteria (0.18 µg biomass) divided into 1000 spots in the pore space. We put a mass of DOM of 0. 2895 mgC in the pore space corresponding to the concentration of 0.13 mgC/g soil, which is the same as the concentration used in the experiments described in Monga et al. (2014). We use the molecular diffusion coefficient of DOM in water of 6.73 10$^{-6}$ cm$^2$ s$^{-1}$. Figure 21 shows the good similarity between the biological dynamics curves provided by the present approach with the ones of LBM and Mosaic models. The diffusive overall conductance coefficient was set to 0.35 according to the calibration phase. We point out that it is remarkable that such different approaches can give so close results. On regular PC, the computing time (CPU) for LBM approach was several days, the one of Mosaic ball based approach was 20 minutes, the one of the present method was 1 minute.

For datasets 2 and 3, we made simply the comparison with MOSAIC (balls network). The biological parameters and the initial masses of micro-organisms and DOM were the same than for dataset 1. We also put homogeneously, in the sense of the concentration in the pores, the Dissolved Organic Matter (0. 2895 mgC). We put two spots of micro-organisms in the two biggest balls for the Mosaic ball based method. The radiuses of the balls were 30 and 27. These balls correspond roughly to the area of the 10 biggest regions when using the curvilinear skeleton based approach. Therefore, we put the same biomass (0.18 µg) into the 10 biggest regions whose total volume is roughly the same than the two biggest balls. Indeed, the comparison of geometrical modelling methods using different volume primitives is more complex than it can appear at first blush. We performed microbial decomposition simulation for real duration of 30 days. We put the same constant value for diffusive overall conductance coefficient (0.35) than for dataset 1. Figure 22 shows that the biological dynamic curves are almost the same for the two methods showing the pertinence of the two pore space geometrical modelling approaches. It shows also that the diffusive overall conductance coefficient does not need (at least for these datasets) to be updated when changing datasets.

In conclusion, the simulation results using the curvilinear skeleton based pore space representation fit well with the ones of other approaches. The benefit is mainly in terms of the computational cost which is drastically reduced. These results illustrate also the pertinence of the pore space representation from curvilinear skeleton.

# 6    Conclusion

This paper deals with the geometrical modeling of the pore space from 3D micro CT (Computed Tomography) images of soil samples, based on the curvilinear skeleton. Indeed, many works use the curvilinear skeleton to represent complex volume shapes within other application domains: medical, material sciences, chemical… In the area of porous media analysis, previous studies focused on rocks micro-structures modelling (carbonate rocks,…). As far as we know, this paper is among the first ones coping with the use of curvilinear skeleton to represent soil samples micro-structures.

The principle of our method consists in segmenting the curvilinear skeleton into simple branches, and afterward in attaching to each simple branch a connected set of points. The result is a partition of the pore space which can be represented by an attributed relational graph. In this graph, each node corresponds to a pore and each arc to an adjacency between a couple of pores.

We show that this geometrical representation of the pore space can be used directly to simulate microbial decomposition dynamics including diffusion and transformation processes. We validate the simulation results by comparison with other classical methods. The advantage of the curvilinear skeleton based pore space modelling is four fold. It does not impose specific shapes for a pore as primitive based modelling methods (balls, ellipsoids…). It defines an exact piecewise representation of pore space because providing a partition of the pore space (no pore space is lost). When implementing diffusion processes, it yields using the exact values of the contact surfaces. The number of nodes (pores) is much less than the balls network used in a previous work. The drawback, compared with geometrical primitives based modelling, is that each pore is defined by a set of connected voxels with no explicit geometric properties.

In further works, we will investigate how to implement hybrid geometric modelling algorithms using both curvilinear/surface skeletons and geometric primitives.

# 7 Acknowledgments

The research described was made possible through CNRST grant I-Maroc (APRD program).

# 8 Code availability

The corresponding code can be downloaded at

https://github.com/belghali/curvilinearskeleton

# 9 Figures

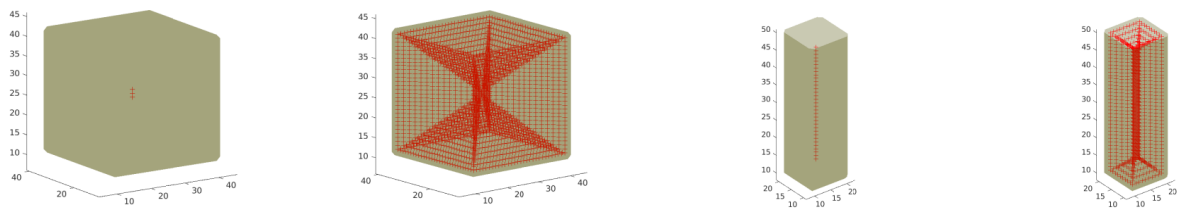

Figure 1 : Curve skeleton and surface skeleton of both a cubic and a parallelepiped shapes.

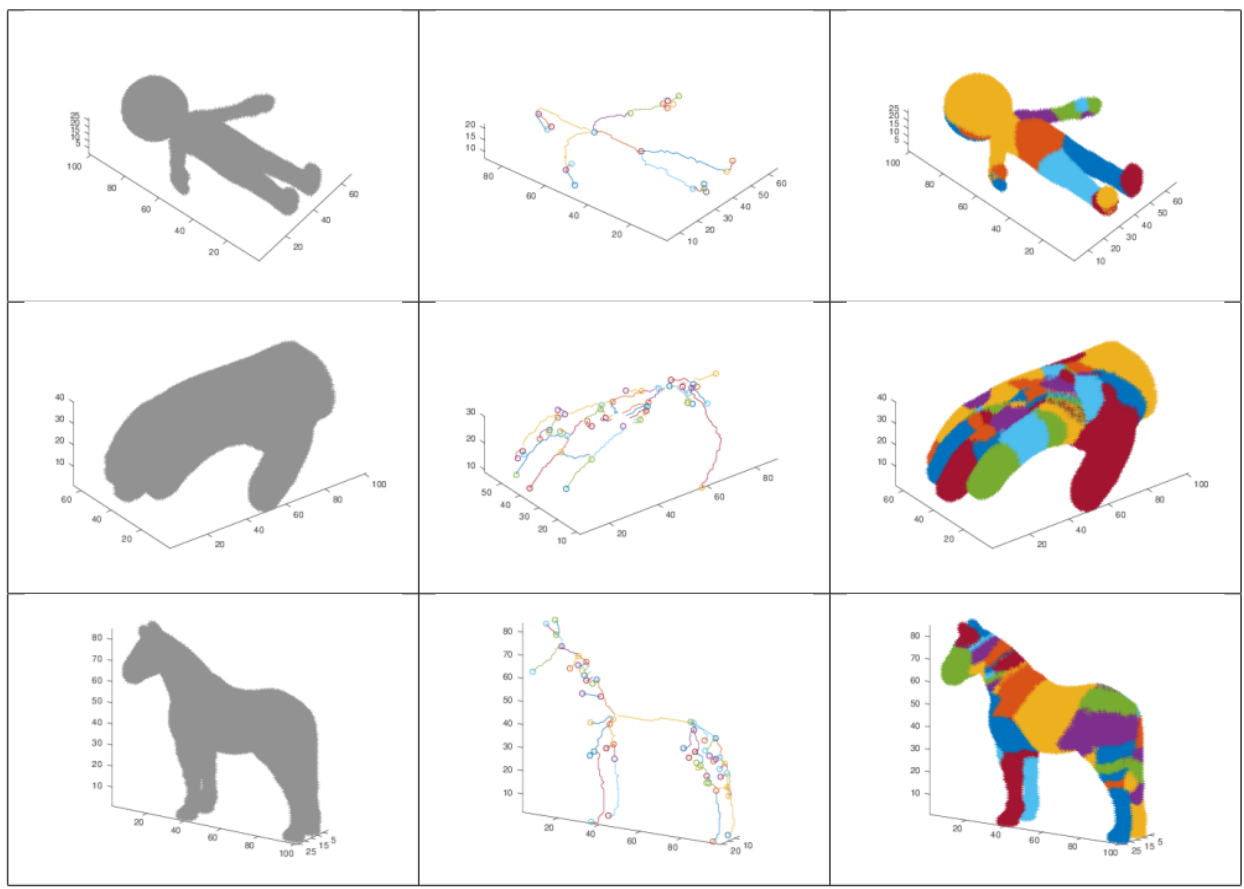

Figure 2: Each row corresponds to a familiar Shape ; from left to right : picture of the voxelized shape, curve skeleton, branch-based segmentation of the shape (each (1D manifold) branch of the curve skeleton is attached to a (3D) region) ; calculation was done with our Matlab implementation ; 3D Models are coming from free3D database Free3d (2019).

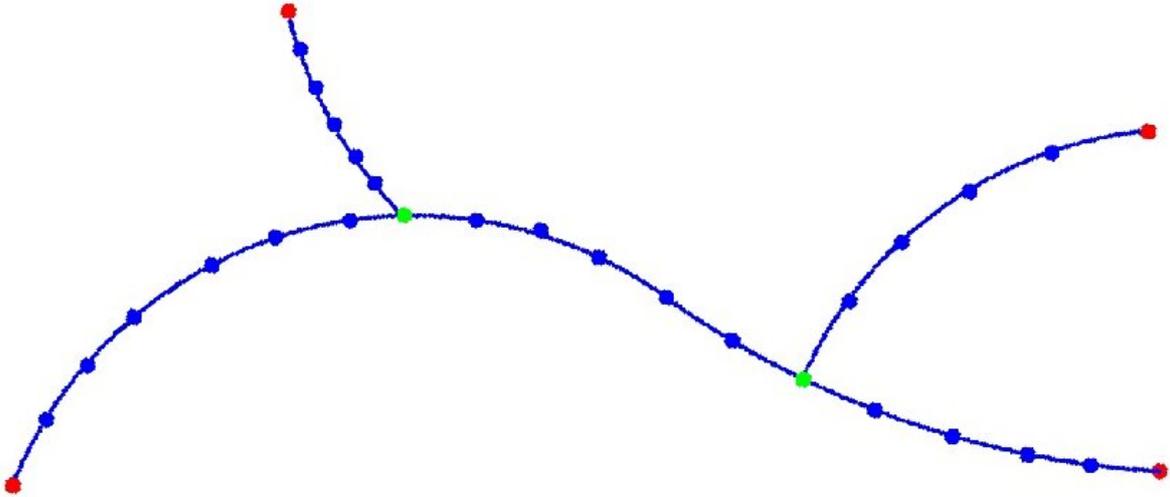

Figure 3: Representation of different kinds of nodes in skeleton graph. Interior nodes are green. Ending nodes are red. Simple nodes are blue.

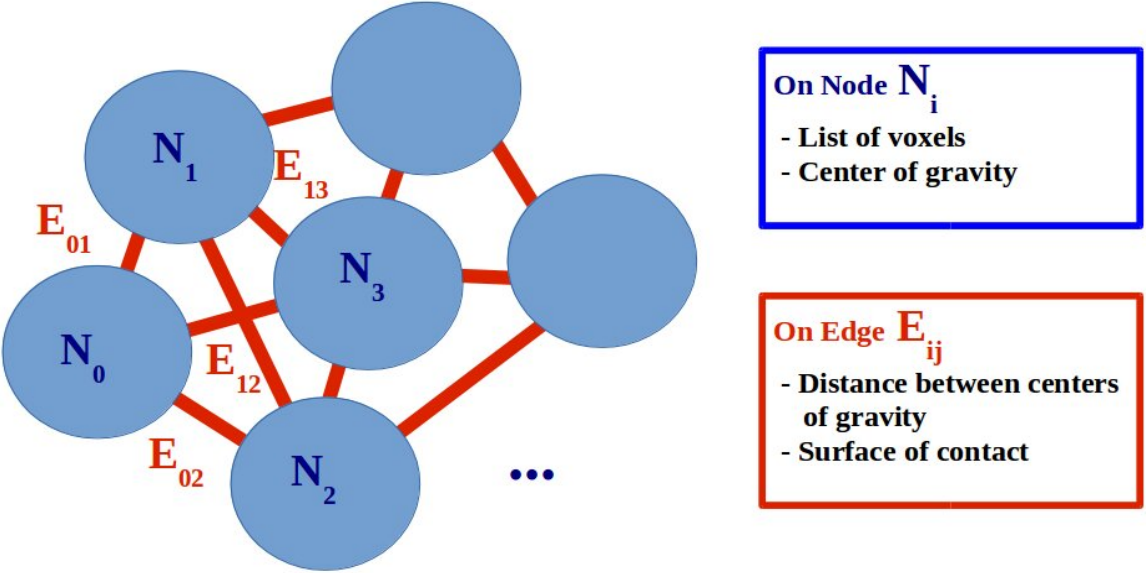

Figure 4: Information stored within the graph of regions after segmentation from curve skeleton. A node is attached to a region and an edge is created between two neighboring regions.

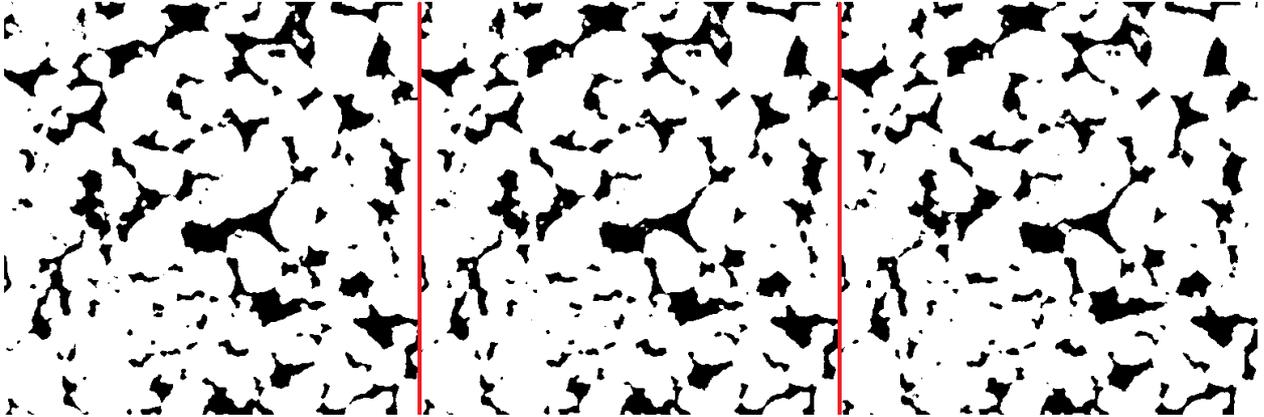
Figure 5 : Three successive cross sections of sandy loam soil ($512 \times 512 \times 512$ ; $24\mu m \times 24\mu m \times 24\mu m$) ; each cross section is delimited by the red line segment, pore space voxels are colored in black (17% of the whole image) ; the water saturation is 100%

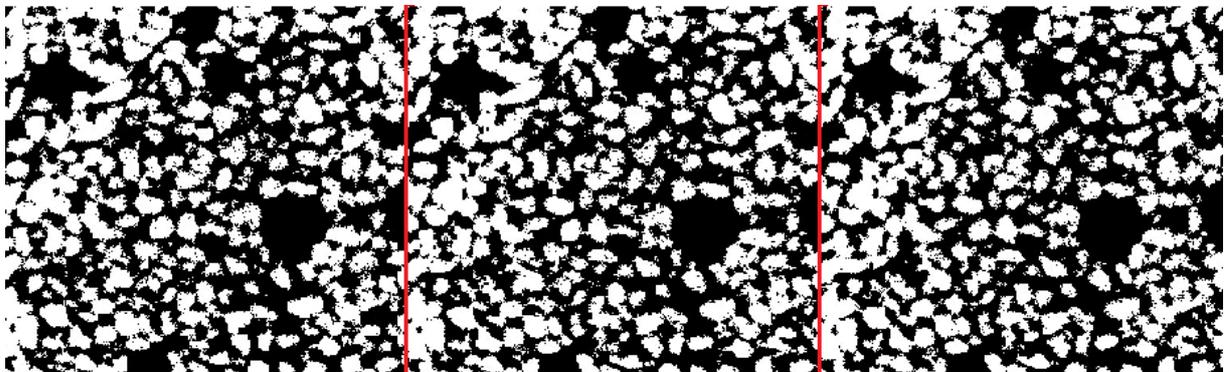
Figure 6 : Three successive cross sections of Fontainebleau sand (high) ($512 \times 512 \times 512$ ; $4.5\mu m \times 4.5\mu m \times 4.5\mu m$) ; each cross section is delimited by the red line segment, pore space voxels are colored in black (55% of the whole image).

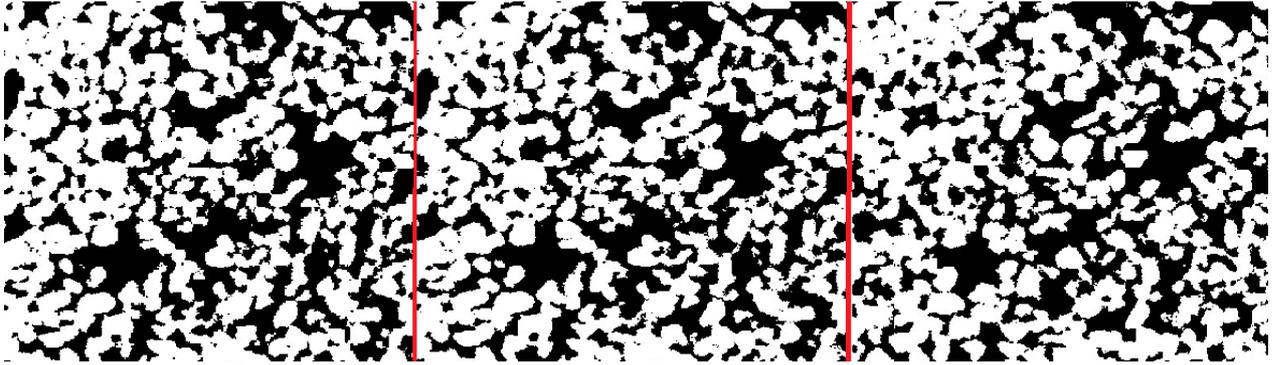

Figure 7 : Three successive cross sections of Fontainebleau sand (low)
($512 \times 512 \times 512$ ; $4.5 \mu m \times 4.5 \mu m \times 4.5 \mu m$) ; each cross section is delimited by the red line segment, pore space voxels are colored in black (42% of the whole image).

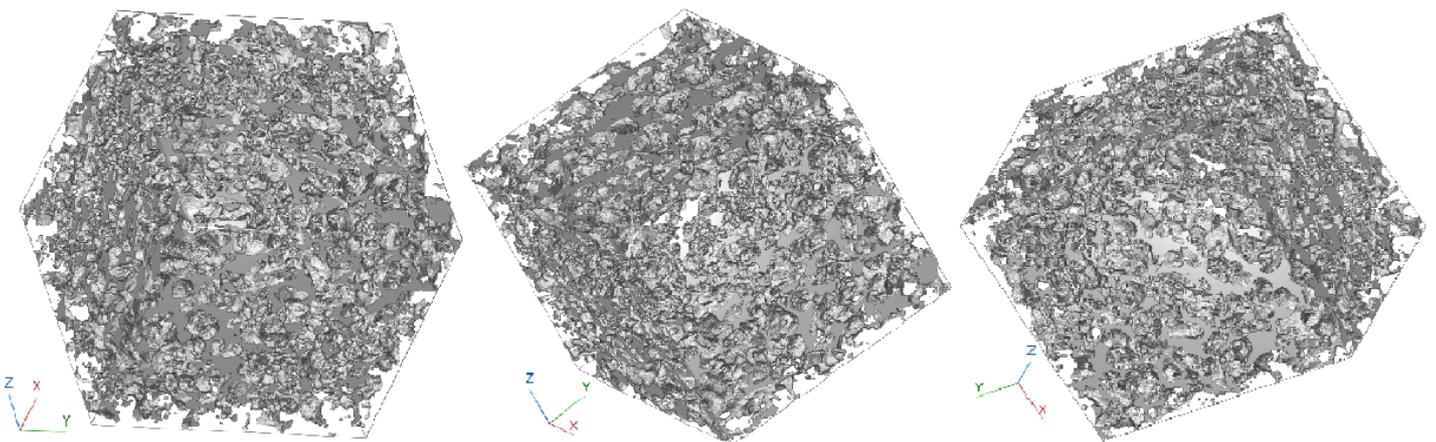

Figure 8 : Perspective views of the pore space (sandy loam soil) ; pore space is colored in grey.

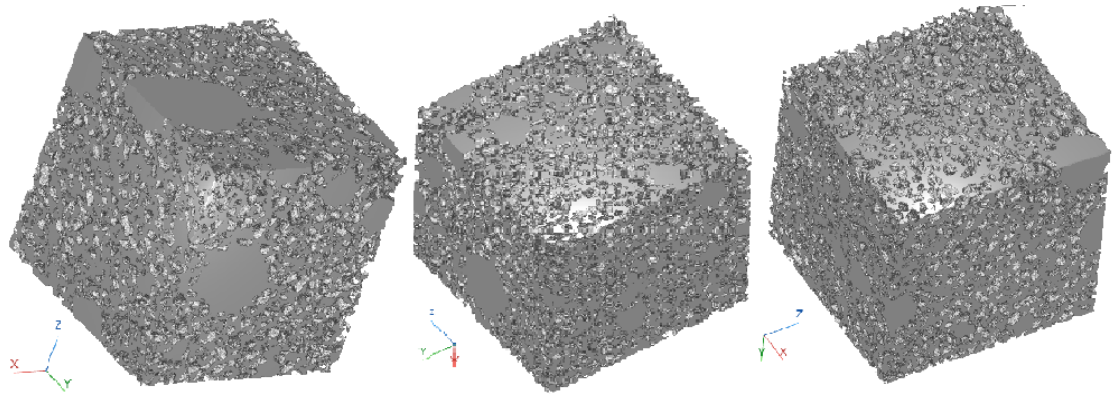

Figure 9 : Perspective views of the pore space (Fontainebleau sand high porosity) ; pore space is colored in grey (dataset 2)

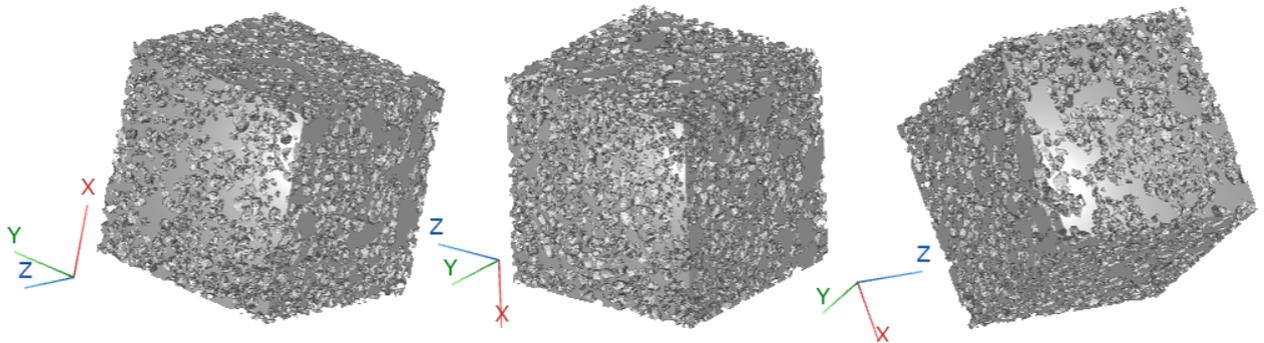

Figure 10 : Perspective views of the pore space (Fontainebleau sand low porosity) ; pore space is colored in grey (dataset 3)

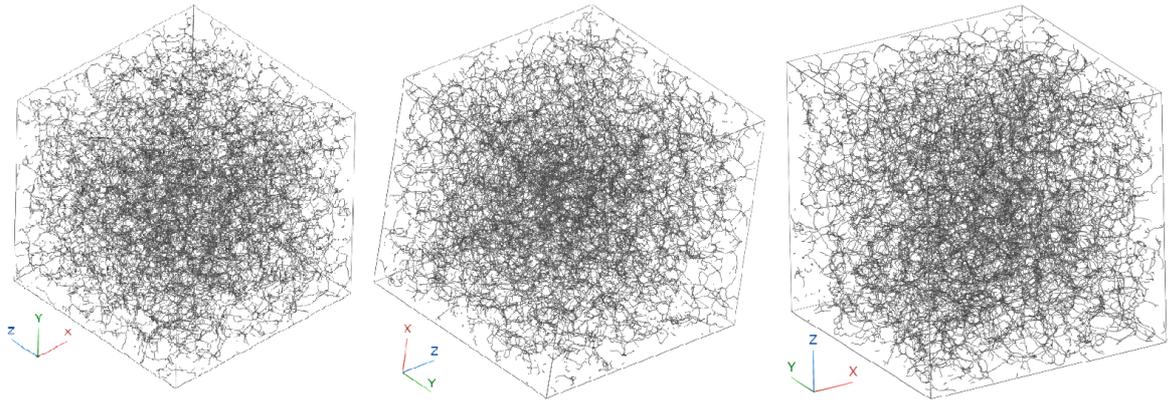

Figure 11 : Perspective views of the curvilinear skeleton (209992 points) of sandy loam soil (dataset 1).

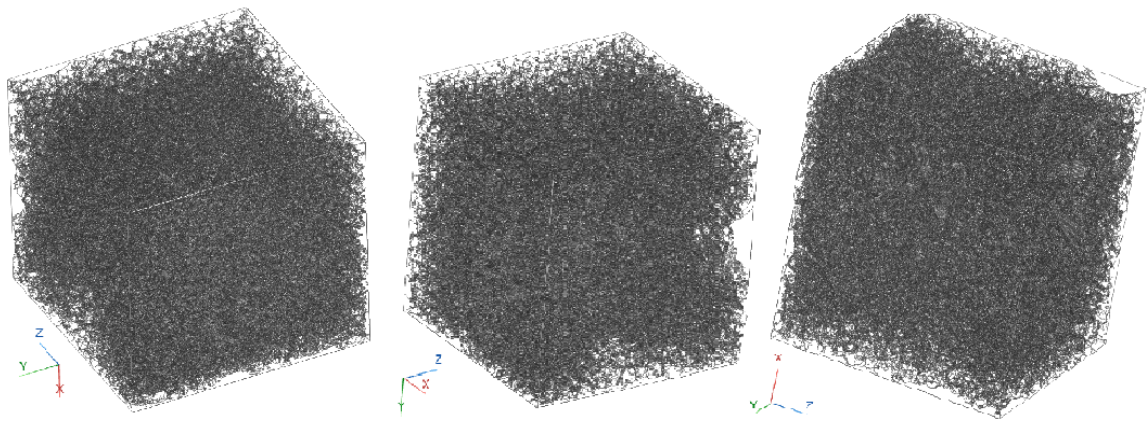

Figure 12 : Perspective views of the curvilinear skeleton (3688771 points) of Fontainebleau sand high porosity (dataset 2).

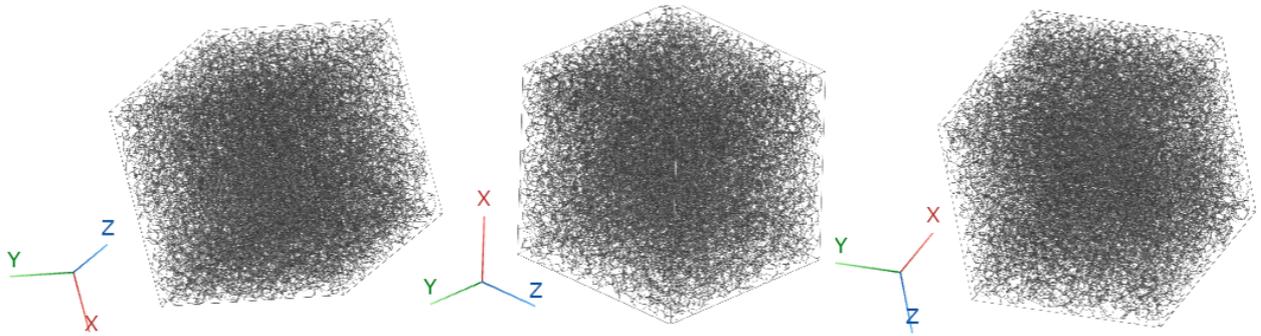

Figure 13 : Perspective views of the curvilinear skeleton (659029 points) of Fontainebleau sand low porosity (dataset 3).

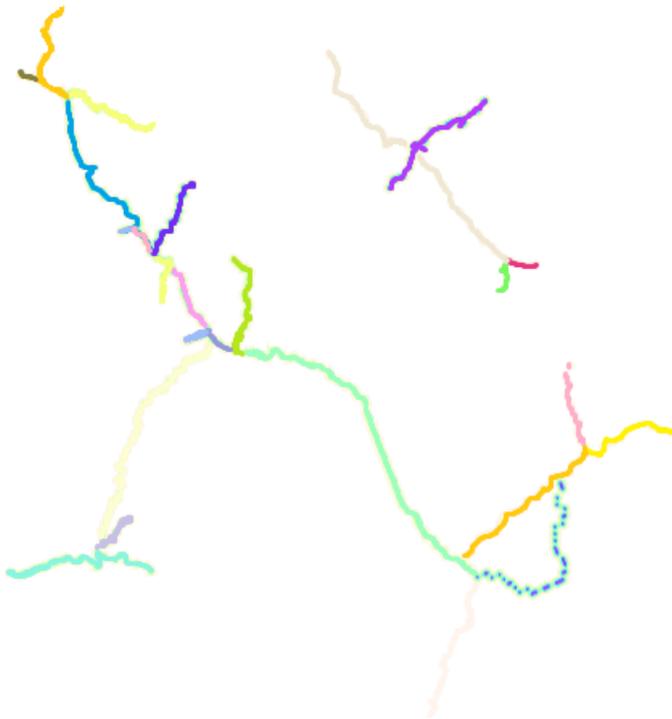

Figure 14: Segmentation of the curvilinear skeleton into simple branches for dataset 1 ; zoom on some simple branches (18508 simple branches).

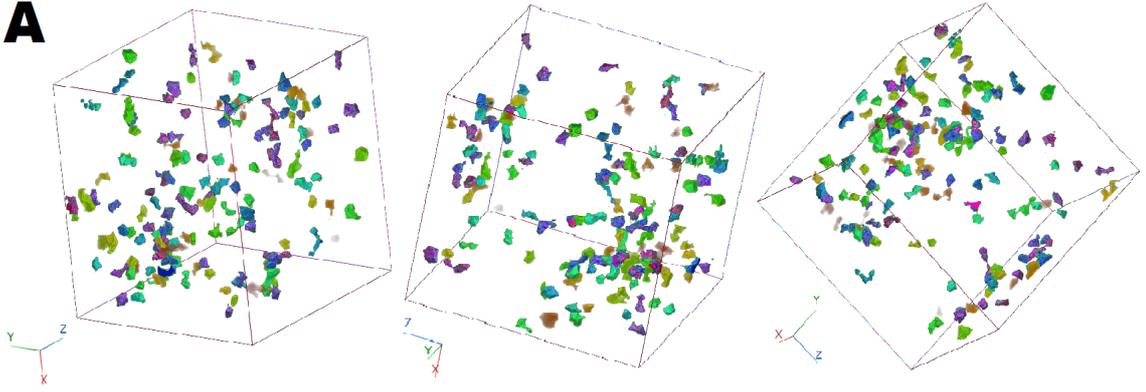

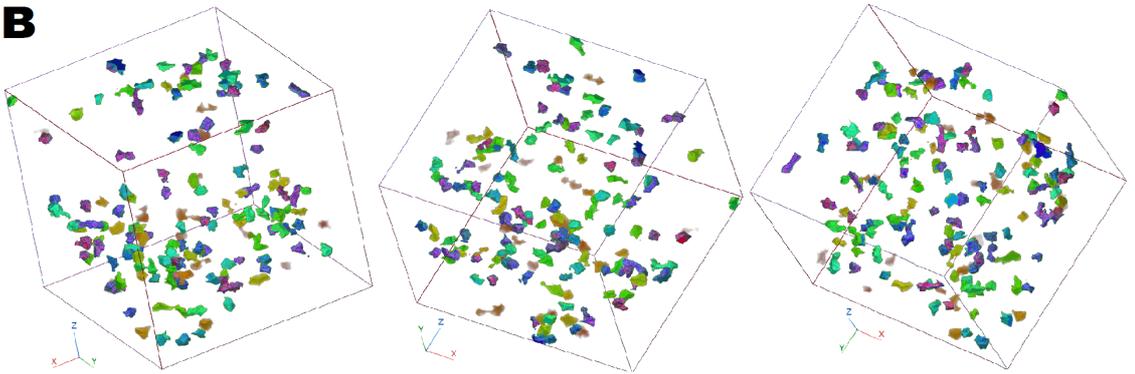

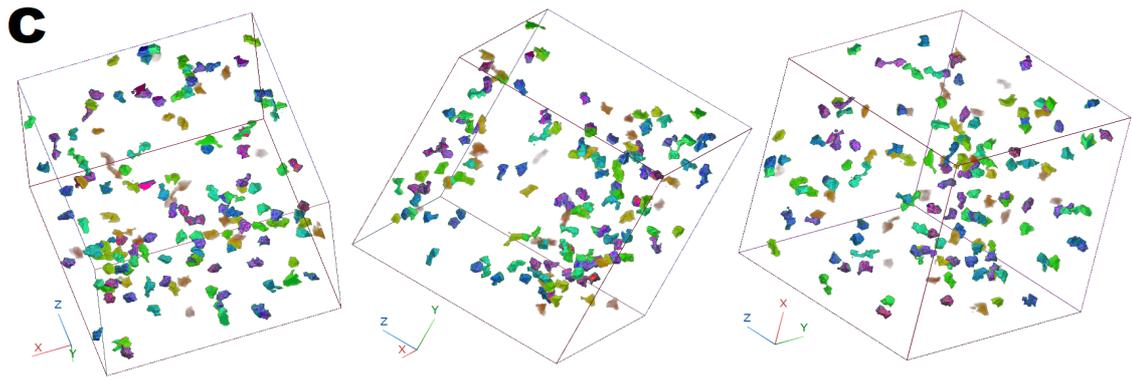

Figure 15: Perspective views three random sets of 150 regions (pores) corresponding to simple branches (dataset1). Each pore is colored with a single color.

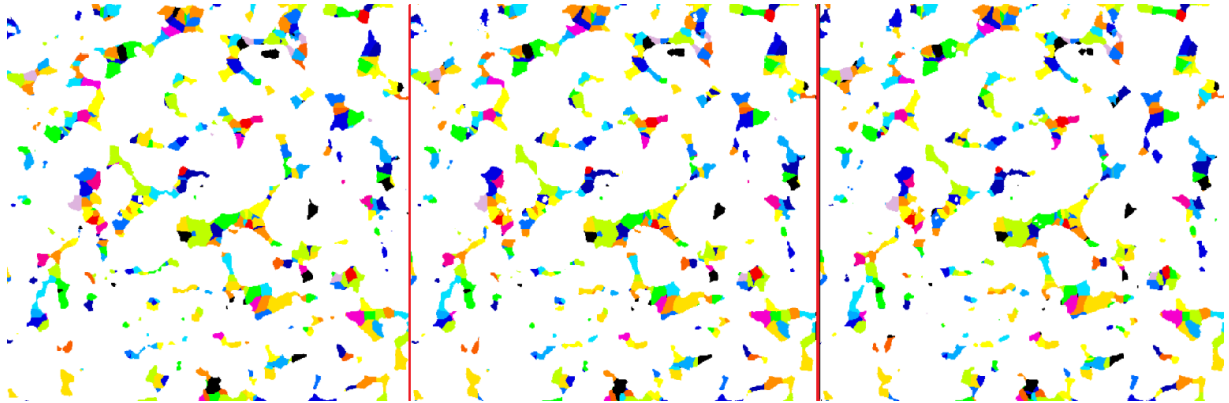

Figure 16: Pore space partition for data set 1 ; cross sections corresponding to figure 5 where each region (pore) is colored with a specific color (label image).

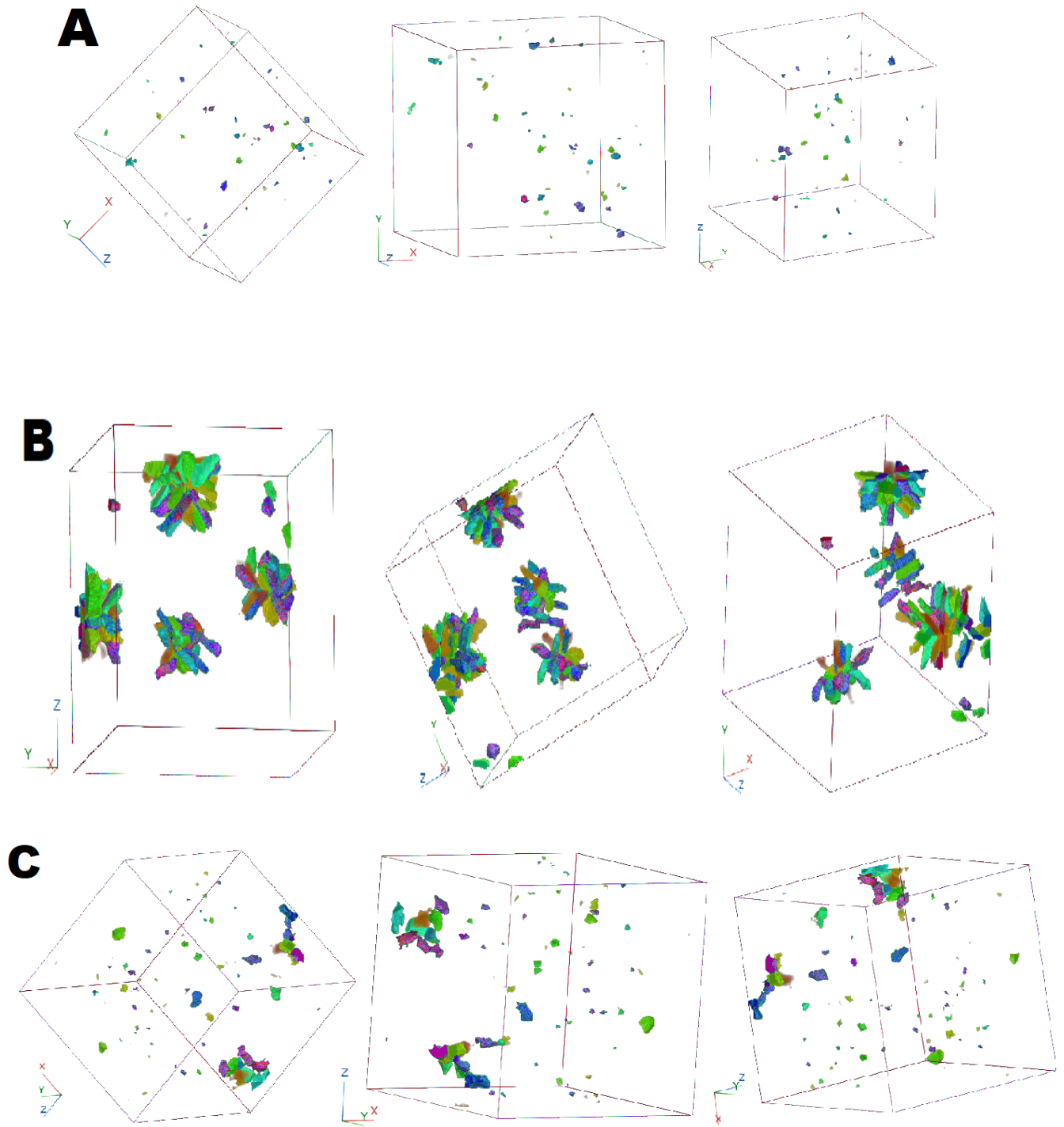

Figure 17 : Perspective views of pores(sand high porosity, dataset 2). **A and C:** sets of random 150 regions (pores) corresponding to simple branches. Each pore is colored with a single color. **B:** the 150 biggest pores

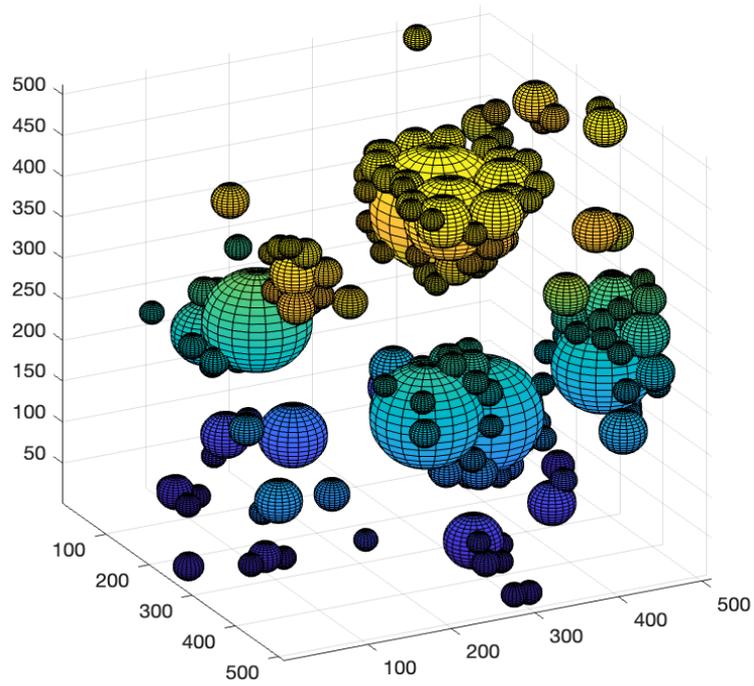

Figure 18: Perspective view of the optimal ball network (dataset 2) ; the balls whose radius is bigger than 13 are displayed (185 balls) ; we notice the good concordance with the pores provided by curvilinear skeleton based method.

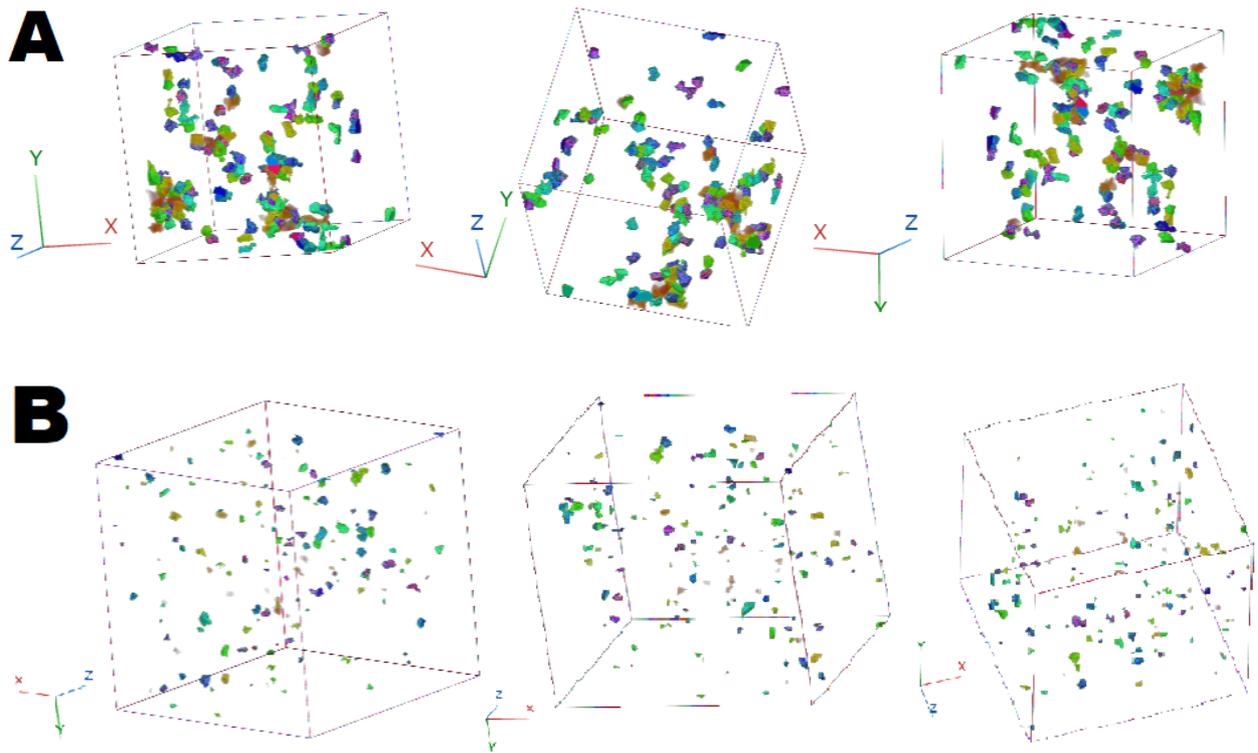

Figure 19: Perspective views of pores(sand low porosity, dataset 3). **B:** sets of random 150 regions (pores) corresponding to simple branches. Each pore is colored with a single color. **A:** the 150 biggest pores

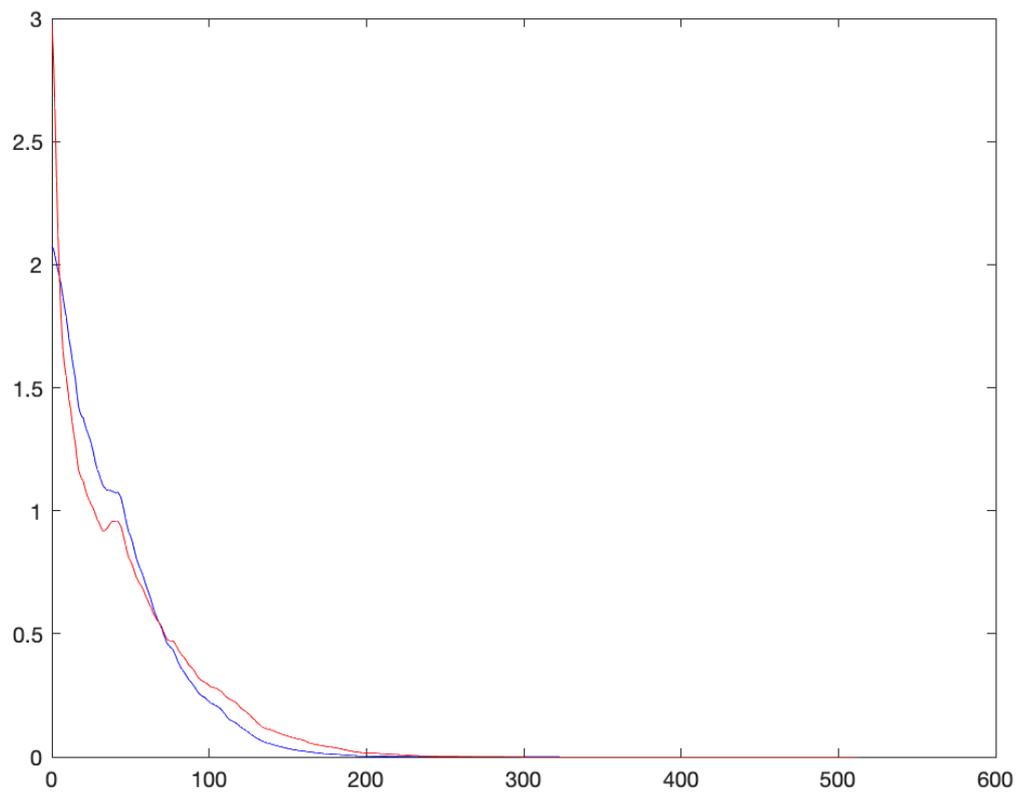

Figure 20: Calibration of diffusion for data set 1. X-axis displays the numbers of the planes (300 planes in total) ; Y-axis displays the total mass of organic matter within the plane ; the real simulation time is 1.783 hours ; we injected 592,7593 mg of carbon within the two first planes. The optimal value of the diffusive overvall conductance coefficient is 0.35 (we fixed it at a constant value) with an intercorrelation of 0.9818.

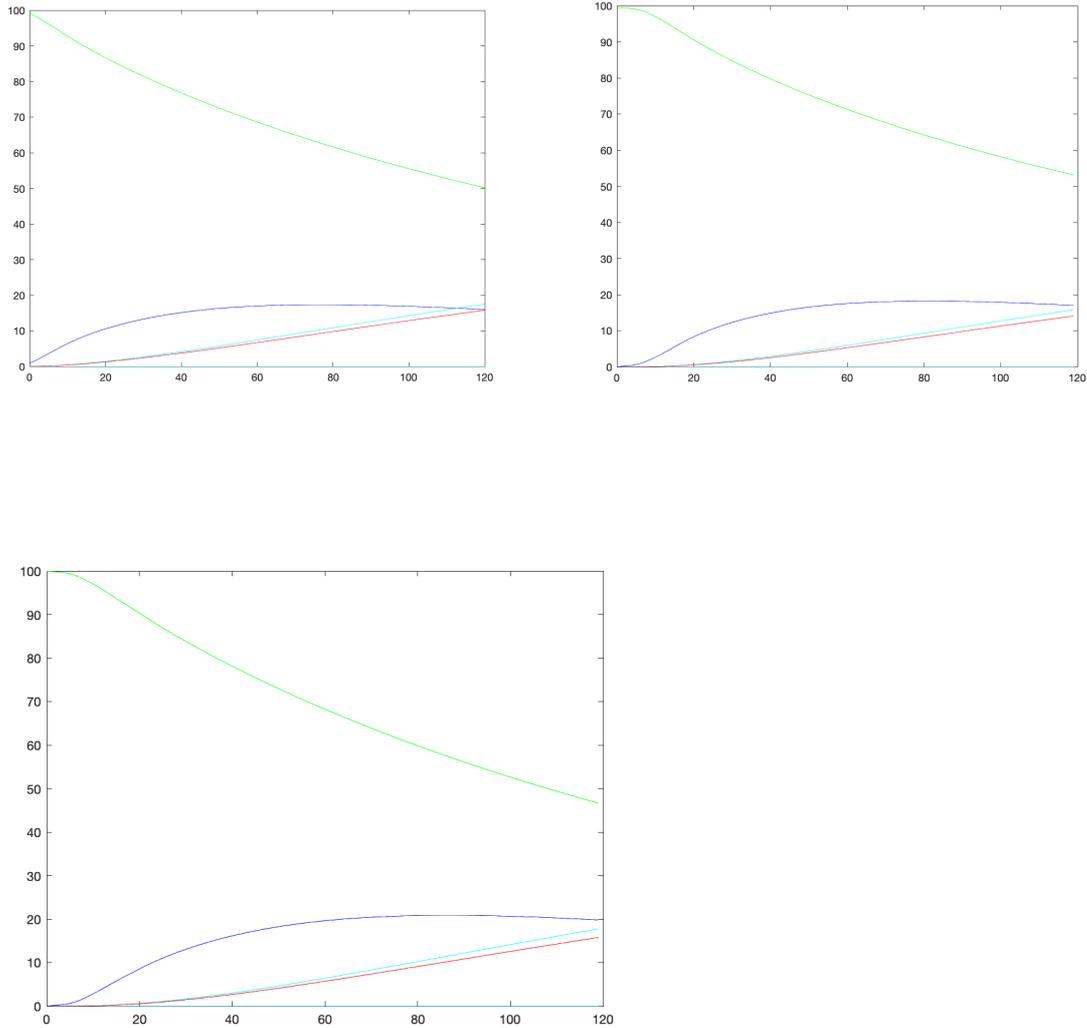

Figure 21: Comparison of microbial decomposition simulation for dataset 1. X-axis displays the time expressed in hours (total simulation time: 5 days). Y axis displays the percentage of the total initial masses. Red, dark blue, green, light blue are respectively attached to CO2, biomass, dissolved organic matter (DOM), Solid Organic Matter (SOM). According to the calibration stage (see figure 19), the diffusive overall conductance coefficient was set to a constant value of 0.35. Up left: curves obtained using LBM (several weeks CPU time) ; up right: curves obtained using Mosaic (ball based model, 20 minutes CPU time), bottom: curves obtained using regions from curvilinear skeleton (1 minute CPU time).

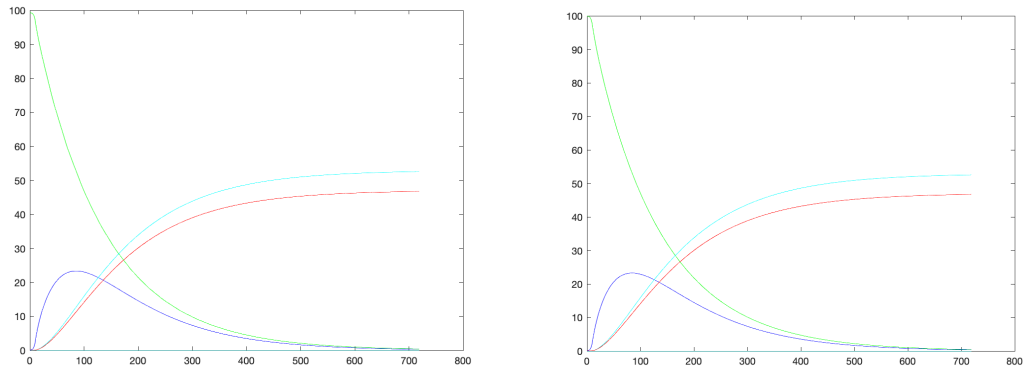

Figure 22 : Comparison of microbial decomposition simulation for dataset 2. Same format than in figure 20. Left: curves obtained using Mosaic ball based model (CPU time: 10h). Right: curves obtained using curvilinear skeleton based model (CPU time: 1h). We set the diffusive conductance coefficient to the same constant value than for dataset 1 (0.35). The real simulation time is one month.

## 10 Acknowledgements